\documentclass[11pt,a4paper]{article}
\usepackage{times,latexsym}
\usepackage{url}
\usepackage[T1]{fontenc}

\usepackage[acceptedWithA]{tacl2018v2}

\usepackage{xspace,mfirstuc,tabulary}

\newif\iftaclinstructions
\taclinstructionsfalse 
\iftaclinstructions
\renewcommand{\confidential}{}
\renewcommand{\anonsubtext}{(No author info supplied here, for consistency with
TACL-submission anonymization requirements)}
\newcommand{\instr}
\fi

\iftaclpubformat 

\else

\fi

\usepackage{amsmath,amsfonts,bm}
\usepackage{microtype}
\usepackage{graphicx}
\usepackage{booktabs} 
\usepackage{multirow}
\usepackage{microtype}
\usepackage{subcaption}
\usepackage{marginnote}
\usepackage{bbm}
\usepackage{float}
\usepackage[shortlabels]{enumitem}
\usepackage{tikz}
\usepackage{ifthen}
\usepackage{stmaryrd}
\usepackage{wrapfig}
\usepackage{caption}
\usepackage{soul}
\usepackage{pifont}
\usepackage{nicefrac}
\usepackage{adjustbox}
\usepackage{cancel}

\usepackage{inconsolata}

\newif\ifcomments
\commentstrue
\ifcomments
    \providecommand\matt[1]{\textcolor{teal}{[Matt: #1]}}
    \providecommand\nitish[1]{\textcolor{violet}{[NG: {#1}]}}
    \providecommand\dr[1]{\textcolor{olive}{[DR: #1]}}
    \providecommand\sameer[1]{\textcolor{purple}{[Sameer: #1]}}
    \providecommand\dd[1]{\textcolor{blue}{[DD: #1]}}
        
\else
    \providecommand{\matt}[1]{}
    \providecommand{\nitish}[1]{}
    \providecommand{\dr}[1]{}
    \providecommand{\sameer}[1]{}
    \providecommand{\dd}[1]{}
\fi

\newcommand{\qdmr}{\textsc{Break}}
\newcommand{\drop}{\textsc{DROP}}
\newcommand{\nmn}{\textsc{NMN}}
\newcommand{\mtmsn}{\textsc{MTMSN}}

\newcommand{\pairedloss}{\mathcal{L}_{\text{paired}}}

\newcommand{\foundloss}{$\mathcal{L}_{\text{paired, found}}$}
\newcommand{\temploss}{$\mathcal{L}_{\text{paired, temp}}$}
\newcommand{\qgenloss}{$\mathcal{L}_{\text{paired, qgen}}$}
\newcommand{\allloss}{$\mathcal{L}_{\text{paired, all}}$}

\newcommand{\module}[1]{\texttt{#1}}
\newcommand{\question}[1]{\emph{#1}}
\newcommand{\modarg}[1]{[\emph{#1}]}

\newcommand{\find}[1]{\texttt{find}}
\newcommand{\spans}[1]{\texttt{spans}}
\newcommand{\filter}[1]{\texttt{filter}}
\newcommand{\countmod}[1]{\texttt{count}}
\newcommand{\project}[1]{\texttt{project}}
\newcommand{\numadd}[1]{\texttt{num-add}}
\newcommand{\numdiff}[1]{\texttt{num-diff}}
\newcommand{\findnum}[1]{\texttt{find-num}}
\newcommand{\finddate}[1]{\texttt{find-date}}
\newcommand{\timediff}[1]{\texttt{time-diff}}
\newcommand{\numcomparelt}[1]{\texttt{num-compare-lt}}
\newcommand{\numcomparegt}[1]{\texttt{num-compare-gt}}
\newcommand{\datecomparelt}[1]{\texttt{date-compare-lt}}
\newcommand{\datecomparegt}[1]{\texttt{date-compare-gt}}
\newcommand{\findmaxnum}[1]{\texttt{find-max-num}}
\newcommand{\findminnum}[1]{\texttt{find-min-num}}

\newcommand{\denotation}[1]{\llbracket #1 \rrbracket}

\title{
Paired Examples as Indirect Supervision in Latent Decision Models
}

\author{
 Nitish Gupta$^{1}$ \quad Sameer Singh$^{2}$ \quad Matt Gardner$^{3}$ \quad Dan Roth$^{1}$ \\
 $^{1}$University of Pennsylvania, $^{2}$University of California, Irvine, $^{3}$Allen Institute for AI \\
 {\sf \{nitishg, danroth\}@seas.upenn.edu, sameer@uci.edu, mattg@allenai.org}
}

\date{}

\begin{document}
\maketitle
\begin{abstract}
Compositional, structured models are appealing because they explicitly decompose problems and provide interpretable intermediate outputs that give confidence that the model is not simply latching onto data artifacts. Learning these models is challenging, however, because end-task supervision only provides a weak indirect signal on what values the latent decisions should take. This often results in the model failing to learn to perform the intermediate tasks correctly. In this work, we introduce a way to leverage \emph{paired examples} that provide stronger cues for learning latent decisions. When two related training examples share internal substructure, we add an additional training objective to encourage consistency between their latent decisions. Such an objective does not require external supervision for the values of the latent output, or even the end task, yet provides an additional training signal to that provided by individual training examples themselves. We apply our method to improve compositional question answering using neural module networks on the DROP dataset. We explore three ways to acquire paired questions in DROP: (a)~discovering naturally occurring paired examples within the dataset, (b)~constructing paired examples using templates, and (c)~generating paired examples using a question generation model. We empirically demonstrate that our proposed approach improves both in- and out-of-distribution generalization and leads to correct latent decision predictions.
\end{abstract}

\section{Introduction}
Developing models that are capable of reasoning about complex real-world problems is challenging.  It involves decomposing the problem into sub-tasks, making intermediate decisions, and combining them to make the final prediction.
While many approaches develop black-box models to solve such problems, we focus on compositional structured models as they provide a level of explanation for their predictions via interpretable latent decisions, and should, at least in theory, generalize better in compositional reasoning scenarios.
For example, to answer \question{How many field goals were scored in the first half?} against a passage containing a football-game summary, a neural module network~\citep[NMN;][]{NMN16} would first ground the set of \emph{field goals} mentioned in the passage, then filter this set to the ones scored \emph{in the first half}, and then return the size of the resulting set as the answer.

Learning such models using just the end-task supervision is difficult, since the decision boundary that the model is trying to learn is complex, and the lack of any supervision for the latent decisions provides only a weak training signal.
Moreover, the presence of dataset artifacts~\citep[\emph{among others}]{Lai2014IllinoisLHAD, Gururangan2018AnnotationAI, Min2019CompositionalQD}, and degeneracy in the model, where incorrect latent decisions can still lead to the correct output, further complicates learning. As a result, models often fail to predict meaningful intermediate outputs and instead end up fitting to dataset quirks, thus hurting generalization~\citep{subramanian-etal-2020-obtaining}.

We propose a method to leverage related training examples to provide an indirect supervision to these intermediate decisions.  
Our method is based on the intuition that related examples involve similar sub-tasks; hence, we can use an objective on the outputs of these sub-tasks to provide an additional training signal. 
Concretely, we use \emph{paired examples}---instances that share internal substructure---and apply an additional training objective relating the outputs from the shared substructures resulting from partial model execution.
Using this objective does not require supervision for the output of the shared substructure, or even the end-task of the paired example. This additional training objective imposes weak constraints on the intermediate outputs using related examples and provides the model with a richer training signal than what is provided by a single example.
For example, \question{What was the shortest field goal?} shares the substructure of finding all \emph{field goals} with \question{How many field goals were scored?}.
For this \emph{paired example}, our proposed objective would enforce that the output of this latent decision for the two questions is the same.

We demonstrate the benefits of our paired training objective using a textual-NMN~\citep{GuptaNMN20} designed to answer complex compositional questions on \drop{}~\citep{DROPDua17}, a dataset requiring natural language and symbolic reasoning against a paragraph of text. 
While there can be many ways of acquiring paired examples, we explore three directions. 
First, we show how naturally occurring paired questions can be automatically found from within the dataset.
Further, since our method does not require end-task supervision for the paired example, one can also use data augmentation techniques to acquire paired questions without requiring additional annotation. We show how paired questions can be constructed using simple templates, and how a 
question generation model can be used to generate paired questions.

We empirically show that our paired training objective leads to overall performance improvement of the NMN model.  While each kind of paired data acquisition leads to improved performance, 
combining paired examples from all techniques leads to the best performance~(\S\ref{ssec:exp:iid}).
We quantitatively show that using this paired objective results in significant improvement in predicting the correct latent decisions~(\S\ref{ssec:exp:faithful}),
and thus demonstrate that the model's performance is improving \emph{for the right reasons}.
Finally, we show that the proposed approach leads to better \emph{compositional generalization} to out-of-distribution examples~(\S\ref{ssec:exp:compgen}).
Our results show that we achieve the stated promise of latent decision models: 
an interpretable model that naturally encodes compositional reasoning and uses its modular architecture for better generalization.

\begin{figure*}[t]
\setlength\abovecaptionskip{7pt}
\setlength\belowcaptionskip{-5pt}
\centering
\includegraphics[width=\textwidth]{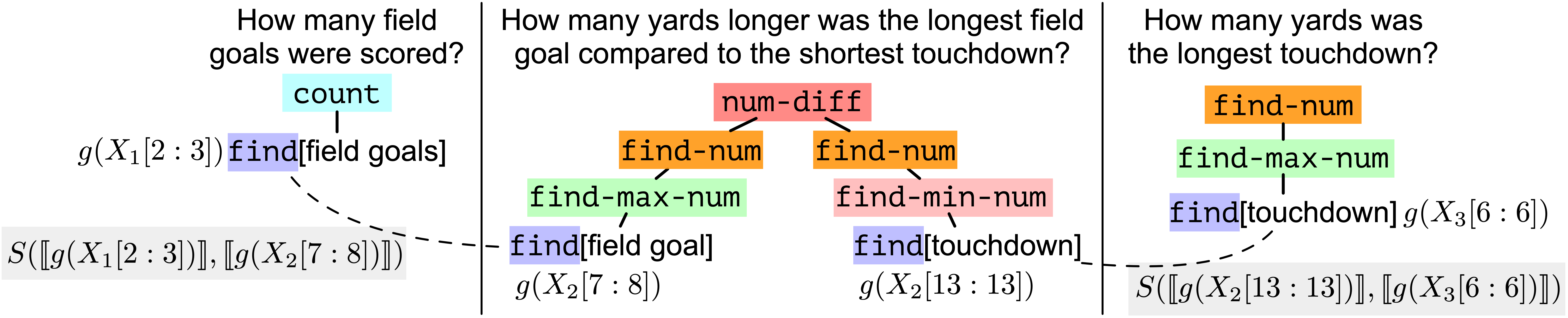}
\caption{\textbf{Proposed paired objective:}
For training examples that share substructure, we propose an additional training objective relating their latent decisions; $S$ in the shaded gray area. In this figure, $g(X_{i}[m:n])$ $=$ $g(\textsc{BERT}(x_{i}, p)[m:n])$, where $\textsc{BERT}(x_{i}, p)$ is the contextualized representation of $x_{i}$-th question/passage, and $[m:n]$ is its slice for the $m$ through $n$ token. $g$ = \find{} in all cases. See \S\ref{sec:paired-nmn} for details. Here, since the outputs of the shared substructures should be the same, $S$ would encourage \emph{equality} between them.
}
\label{fig:overview}
\end{figure*}

\section{Paired Examples as Indirect Supervision for Latent Decisions} 
\label{sec:paired-loss}
We focus on structured compositional models for reasoning that perform an explicit problem decomposition and predict interpretable latent decisions that are composed to predict the final output.
These intermediate outputs are often grounded in real-world phenomena and provide some explanation for the model's predictions.
Such models assume that the structured architecture provides a useful inductive bias for efficient learning. 
For example, for a given input $x$, a model could perform the computation $f(g(x), h(x))$ to predict the output $y$. 
In this paper, we will need to distinguish between a computation tree, and the output of its execution. Therefore, we will use the notation $z$ to denote a computation tree, and 
$\denotation{z}$ to denote the output of its execution.
Hence we can write,
\begin{align}
    y & = \denotation{f(g(x), h(x))}
\end{align}
where $f$, $g$, and $h$ perform the three sub-tasks required for $x$ and
the computations $g(x)$ and $h(x)$ are the intermediate decisions. The actual computation tree would be dependent on the input and the structure of the model.
For example, to answer \question{How many field goals were scored?}, a NMN would perform $f(g(x))$ where $g(x)$ would output the set of \emph{field goals} and $f$ would return the size of this set. 
While we focus on NMNs, other models that have similar structures where our techniques would be applicable include language models with latent variables for coreference~\citep{Ji2017DynamicER}, syntax trees~\citep{Dyer2016RecurrentNN}, or knowledge graphs~\citep{KGLM2019Robert}; checklist-style models that manage coverage over parts of the input~\citep{Kiddon2016GloballyCT}; or any neural model that has some interpretable intermediate decision, including standard attention mechanisms~\citep{Bahdanau2015NeuralMT}.

Typically, the only supervision provided to the model are gold $(x, y^{*})$ pairs, without the outputs of the intermediate decisions ($\denotation{g(x)}$ and $\denotation{h(x)}$ above), from which it is expected to jointly learn the parameters of all of its components.
Such weak supervision is not enough for accurate learning, and the fact that incorrect latent decisions can lead to the correct prediction further complicates learning.
Consequently, models fail to learn to perform these latent tasks correctly and usually end up modeling irrelevant correlations in the data~\citep{Johnson2007WhyDE, subramanian-etal-2020-obtaining}.

In this work, we propose a method to leverage \emph{paired examples}---examples whose one or more latent decisions are related to each other---to provide an indirect supervision to these latent decisions. Consider paired training examples $x_i$ and $x_j$ with the following computation trees:
\begin{align}
    z_i & = f(g(x_i), h(x_i)) \\
    z_j & = f(k(g(x_j)))
\end{align}
These trees share the internal substructure $g(x)$. In such a scenario, we propose an additional training objective 
$S(\denotation{g(x_i)}, \denotation{g(x_j)})$
to enforce consistency of partial model execution for the shared substructure:
\begin{equation}
    \pairedloss = S(\denotation{g(x_i)}, \denotation{g(x_j)})
\label{eq:paired-obj}
\end{equation}
For example, the two questions on the LHS of Figure~\ref{fig:overview} share the intermediate decision of finding the field goals. i.e., their computation trees share the substructure $g(x)$ = \find{}\modarg{field goal}.
In such a case, where the outputs of the intermediate decision should be the same for the paired examples, using a similarity measure for $S$ would enforce equality of the latent outputs $\denotation{g(x)}$.
We will go into the specifics of this example in Section~\ref{sec:paired-nmn}.
By adding this consistency objective, we are able to provide an additional training signal to the latent decision using related examples, and hence indirectly share supervision among multiple training examples.
As a result, we are able to more densely characterize the decision boundary around an instance ($x_i$), by using related instances ($x_j$), than what was possible by using the original instance alone.

To use this consistency objective for $x_i$, we do not require supervision for the latent output $\denotation{g(x_i)}$, nor the gold end-task output $y_j^{*}$ for the paired example $x_j$; we only enforce that the intermediate decisions are consistent.
Additionally, we are not limited to enforcing consistency for a single intermediate decision from a single paired example; if $x_i$ shares an additional substructure $h(x)$ with a paired example $x_k$, we can add an additional term 
$S'(\denotation{h(x_i)}, \denotation{h(x_k)})$ 
to Eq.~\ref{eq:paired-obj}.

Our approach generalizes a few previous methods for learning via paired examples.
For learning to ground tokens to image regions,~\citet{contrastive-gupta2020} enforce contrastive grounding between the original and a negative token; this is equivalent to using an appropriate $S$ in our framework. 
A few approaches~\citep{Minervini2018AdversariallyRN, Li2019ALF, LogicGuidedDA-Asai2020} use an additional objective on model outputs to enforce domain-specific consistency between paired examples; this is a special case of our framework where $S$ is used on the outputs $(y_i, y_j)$, instead of the latent decisions.

Using paired examples for indirect supervision on latent decisions should be broadly applicable to a wide class of models, and our general formulation of this technique is, we believe, novel.  However, the specific application of this method to any particular problem is non-trivial, as work needs to be done to acquire paired data and design a suitable $S$ for the model being studied.  In the rest of this work, we present a case study on text-based neural module networks which we believe is promising enough to motivate further applications of this method in future work.

\section{Training via Paired Examples in Neural Module Networks}
\label{sec:paired-nmn}
We apply our approach to improve question answering using a neural module network~\citep[NMN;][]{NMN16} on the \drop{} dataset~\citep{DROPDua17}. \drop{} contains complex compositional questions against natural language passages describing football games and historical events.

NMN is a model architecture aimed at reasoning about natural language against a given context (text, image, etc.) in a compositional manner.
A NMN maps the input utterance into an executable program representing the compositional reasoning structure required to predict the output. 
The program is composed of learnable modules that are designed to perform atomic reasoning tasks. 
For example, to answer $q =$ \question{How many field goals were scored?}, a NMN would parse it into a program $z =$ \countmod{}(\find{}\modarg{field~goals}).  This program then gets executed by the learnable models to produce $y$, essentially performing $y = \denotation{f(g(q))}$, where $f=\countmod{}$ and $g=\find{}$. 

Given a question $q$, the gold program $z^{*}$, and the correct answer $a^{*}$, maximum likelihood training is used to jointly train the parameters of the modules, as well as a parser that produces the gold program $z^{*}$.
Note that the gold program only supervises the \emph{layout} of the modules ($z$ in the example in above) and not the outputs of the intermediate modules.  That is, we have $a^{*} = \denotation{z^{*}}$, but no supervision on subparts of $z^{*}$, such as $\denotation{g(q)}$.
It is extremely challenging to learn the module parameters correctly in the absence of intermediate module output supervision.
Learning is further complicated by the fact that the space of possible intermediate outputs is quite large and incorrect module output prediction can still lead to the correct answer.
For example, the \find{} module in the question above needs to learn to select the spans describing \emph{field goals} among all possible spans in the passage using just the \emph{count value} as answer supervision.
With no direct supervision for the module outputs, the modules can learn incorrect behavior but still predict the correct answer, effectively memorizing the training data. Such a model would presumably fail to generalize.

\paragraph{Text-NMN}
We work with the Text-NMN of \citet{GuptaNMN20} on a subset of DROP which is annotated with gold programs.
Their model contains \find{}, \filter{}, \project{}, \countmod{}, \findnum{}, \finddate{}, \findmaxnum{}, \findminnum{}, \module{num-compare}, \module{date-compare}, \numadd{}, \numdiff{}, \timediff{}, and \spans{} modules. 
The \find{}, \filter{}, and \project{} modules take as input an additional question string argument.
Each module's output is an attention distribution over the relevant support. E.g. \find{}, \filter{}, \project{} output an attention over passage tokens, \findnum{}, \numadd{} over numbers, \finddate{} over dates, etc. 

Given a question $q$ and passage $p$, BERT is used to compute joint contextualized representations for the (question, passage) combination, $\textsc{BERT}(q, p) \in \mathbb{R}^{(|q|+|p|) \times d}$. 
During execution, the modules that take a question span argument as input (e.g. \find{}) operate on the corresponding slice of this contextualized representation. For example, in $q$ = \question{How many field goals were scored?} with program $z$ = \countmod{}(\find{}\modarg{field goals}), to execute \find{}\modarg{field goals}, the model actually executes \find{}($\textsc{BERT}(q, p)[2:3]$).\footnote{All modules also take the (BERT-encoded) passage $p$ as an implicit argument, as well as additional state extracted from the passage such as which tokens are numbers, which we omit throughout the paper for notational simplicity.} Here the slice $[2:3]$ corresponds to the contextualized representations for the 2nd through the 3rd token (\question{field goals}) of the question.
Refer \citet{GuptaNMN20} for details.

\paragraph{Paired training in NMNs}
We consider a pair of questions whose program trees $z$ share a subtree as paired examples.  A shared subtree implies that a part of the reasoning required to answer the questions is the same.
Since some modules take as input a string argument, we define two subtrees to be equivalent \emph{iff} their structure matches and the string arguments to the modules that require them are \emph{semantically equivalent}.
For example, subtrees \findnum{}(\find{}\modarg{passing touchdowns}) and \findnum{}(\find{}\modarg{touchdown passes}) are equivalent, while they are not the same as \findnum{}(\find{}\modarg{touchdown runs}) (we describe how we detect semantic equivalence in \S\ref{sec:paired-data}).

Consider a question $q_i$ that shares the substructure $g(q)$ with a paired question $q_j$.
Since shared substructures are common program subtrees in our case, we encourage the latent outputs, the outputs $\denotation{g(q)}$ of the subtree, to be equal to enforce consistency.
As the outputs of modules are probability distributions, enforcing consistency amounts to minimizing the KL-divergence between the two outputs.
We therefore maximize the following paired objective from Eq.~\ref{eq:paired-obj},
\begin{multline}
    \pairedloss = - \big(\text{KL}\big[\denotation{g(q_i)}\;\|\;\denotation{g(q_j)}\big] + \\
    \text{KL}\big[\denotation{g(q_j)}\;\|\;\denotation{g(q_i)}\big]\big)
\label{eq:paired-nmn}
\end{multline}
where $S(p_1, p_2) = - (\text{KL}[p_1\|p_2] + \text{KL}[p_2\|p_1])$ is the negative symmetric KL-divergence. 

To understand why such an objective is helpful even though the paired examples share exact subtrees, consider the paired examples on the LHS of Figure~\ref{fig:overview}. The substructure $g(q)$ = \find{}\modarg{field goal} is shared between them. 
Even though the input string argument to \find{} is the same, what gets executed is $g(q_i)$ = \find{}($\textsc{BERT}(x_1, p)[2:3]$) and $g(q_j)$ = \find{}($\textsc{BERT}(x_2, p)[7:8]$), i.e. \find{} gets as input different contextualized representations of \question{field goal} from the two questions.
Due to different inputs, the output of \find{} could be different, which would lead to inconsistent behavior and inefficient learning. 
Our paired objective (Eq.~\ref{eq:paired-nmn}) would encourage that these two outputs are consistent, thereby allowing sharing of supervision across examples. 

\begin{figure*}[t]
\setlength\abovecaptionskip{7pt}
\setlength\belowcaptionskip{-5pt}
\centering
\includegraphics[width=0.95\textwidth]{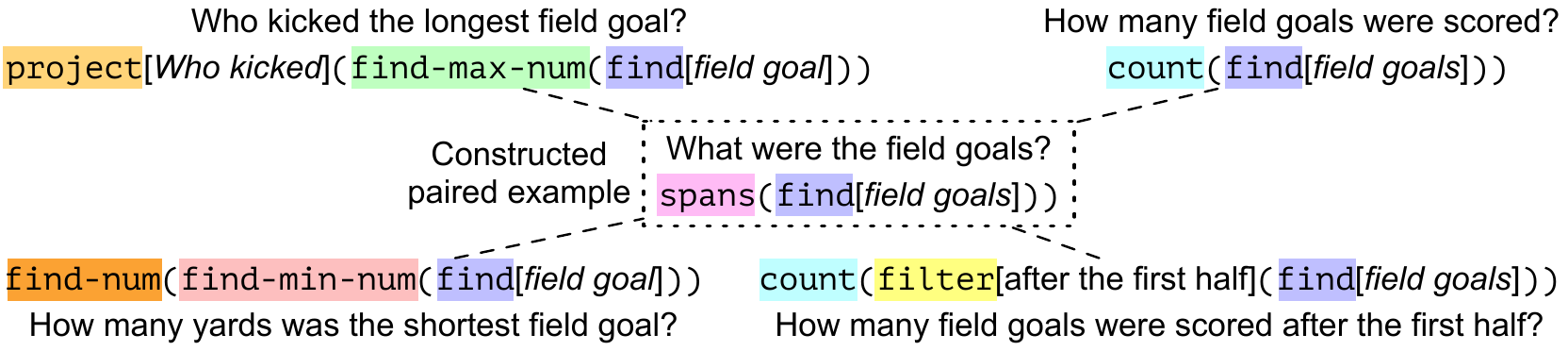}
\caption{\textbf{Templated Construction of Paired Examples}:
Constructed paired examples can help in indirectly enforcing consistency between different training examples~(\S\ref{sssec:construct-paired}).}
\label{fig:construct-paired}
\end{figure*}

\paragraph{Complete Example}
We describe the benefits of training with paired data using an example.
Consider the four questions in the periphery of Figure~\ref{fig:construct-paired}; all of them share the substructure of finding the field goal scoring events. 
However, we find that for the questions requiring the \module{find-\{max/min\}-num} operation, a vanilla NMN directly grounds to the longest/shortest field goal as the \find{} execution.
Due to the use of powerful NNs (i.e., BERT) for contextualized question/passage representations and no constraints on the modules to perform as intended, the model performs the symbolic \emph{min}/\emph{max} operation internally in its parameters.
Such \find{} execution results in non-interpretable behavior, and substantially hurts generalization to the \countmod{} questions.
By enforcing consistency between all the \find{} executions, the model can no longer shortcut the compositional reasoning defined by the programs; this results in correct \find{} outputs and better generalization, as we show in \S\ref{ssec:exp:analysis}.

Note that in this example we do not know the correct answer $a^{*}$ for the constructed question \question{What were the field goals?}, nor do we know the intermediate output $\denotation{\find{}\modarg{field goals}}$.  The \emph{only} additional supervision given to the model is that there is a pairing between substructures in all of these examples, and so the model should be consistent.

\section{Many Ways of Getting Paired Data}
\label{sec:paired-data}
We explore three ways of acquiring paired questions. 
We show how questions that share substructures can be automatically found from within the dataset (\S\ref{ssec:found-data}), and how new paired questions can be constructed using templates (\S\ref{sssec:construct-paired}), or generated using a question-generation model (\S\ref{sssec:generate-paired}).

\subsection{Finding Naturally Occurring Paired Data} 
\label{ssec:found-data}
Any dataset that contains multiple questions against the same context could have questions that query different aspects of the same underlying event or entity. 
These examples can potentially be paired by finding the elements in common between them.  As the \drop{} data that we are using has annotated programs, this process is simplified somewhat in that we can simply find pairs of programs in the training data that share a subtree.
While the subtrees could be of arbitrary size, we limit ourselves to programs that share a leaf \find{} module. Recall that \find{} requires a question string argument, so the challenge of finding paired questions reduces to discovering pairs of \find{} modules in different questions about the same paragraph whose question string arguments are semantically equivalent.
To this end, we use BERTScore~\citep{bert-score} to measure string similarity.

We consider two string arguments to be semantically equivalent if their BERTScore-F1 exceeds a threshold ($0.6$), and if the same entities are mentioned in the arguments. 
This additional constraint allows us to judge that \emph{Jay Feely's field goal} and \emph{Janikowski's field goal} are semantically different, even though they receive a high BERTScore.
This approach would find paired examples like,\\
\question{What term is used to describe \underline{the Yorkist defeat at Ludford Bridge in 1459}?} \\
\question{What happened first: \underline{Yorkist defeat at Ludford Bridge} or widespread pillaging by Queen Margaret?}

\subsection{Paired Data via Augmentation}
\label{ssec:aug-data}
One benefit of our consistency objective (Eq.~\ref{eq:paired-obj}) is that it only requires that the paired example shares substructure.
This allows us to augment training data with new paired questions without knowing their gold answer.
We explore two ways to carry out this augmentation; 
(a) constructing paired questions using templates, and (b) generating paired questions using a question-generation model.

\subsubsection{Templated Construction of Paired Examples}
\label{sssec:construct-paired}

\paragraph{Grounding \find{} event(s)}
Using the question argument from the \find{} module of certain frequently occurring programs, we construct a paired question that aims to ground the mentions of the event queried in the \find{} module.
For example, \question{Who scored the longest touchdown?} would be paired with \question{What were the touchdowns?}.
This templated paired question construction is carried out for,\\
(1) \countmod{}(\find{}\modarg{}) 
(2) \countmod{}(\filter\modarg{}(\find{}\modarg{})) \\ 
(3) \findnum{}(\findmaxnum{}(\find{}\modarg{})) \\
(4) \findnum{}(\findmaxnum{}(\filter{}\modarg{}(\find{}\modarg{})))  \\
(5) \project{}\modarg{}(\findmaxnum{}(\find{}\modarg{})) \\
(6) \project{}\modarg{}(\findmaxnum{}(\filter{}\modarg{}(\find{}\modarg{}))) \\
(7) \datecomparegt{}(\find{}\modarg{}, \find{}\modarg{}) \\
(8) \timediff{}(\find{}\modarg{}, \find{}\modarg{}),
and their versions with \findminnum{} or \datecomparelt{}.

For questions with a program in (1) - (6), we append \question{What were the} to the program's \find{} argument to construct a paired question. We annotate this paired question with the program \spans{}(\find{}\modarg{}), and enforce consistency among the \find{} modules. 
Such a construction allows us to indirectly enforce consistency among multiple related questions via the constructed question; see Figure~\ref{fig:construct-paired}.

For questions with a program in (7) - (8), we append \question{When did the} to the two \find{} modules' arguments and construct two paired questions, one for each \find{} operation. We label the constructions with \finddate{}(\find{}\modarg{}) and enforce consistency among the \find{} modules. 
For example, \question{How many years after the Battle of Rullion Green was the Battle of Drumclog?} would result in the construction of \question{When did the Battle of Rullion Green?} and \question{When did the Battle of Drumclog?}. While this method can lead to ungrammatical questions, it should help in decomposing the two \find{} executions. 

\paragraph{Inverting Superlatives}
For questions with a program in (3) - (6) or its \findminnum{} equivalent, we construct a paired question by replacing the superlative in the question with its antonym (e.g. \emph{largest}~$\rightarrow$~\emph{smallest}) and inverting the min/max module. We enforce consistency among the \find{} modules of the original and the paired question.

\subsubsection{Model-generated Paired Examples}
\label{sssec:generate-paired}
We show how question generation (QG) models~\citep{Du2017LearningTA, Krishna2019GeneratingQH} can be used to generate paired questions.
QG models are seq2seq models that generate a question corresponding to an answer span marked in a passage as input. We follow~\citet{Wang2020AskingAA} and fine-tune a BART model~\citep{Lewis2020BARTDS} on SQuAD~\citep{Rajpurkar2016SQuAD10} to use as a QG model. 

We generate paired questions for non-football passages\footnote{We explain the reason for this in \S\ref{appendix:modgen}} 
in \drop{} by randomly choosing $10$ numbers and dates as answer spans, and generating questions for them.
We assume that the generated questions are SQuAD-like---they query an argument about an event/entity mentioned in text---and label them with the program \findnum{}(\find{}) or \finddate{}(\find{}). We use the whole question apart from the Wh-word as the string argument to \find{}.
We then follow the same procedure as \S\ref{ssec:found-data}---for each of the \find{} module in a \drop{} question's program, we see if a generated question with a \emph{semantically similar} \find{} module exists. If such an augmented question is found, it is used as a paired example for the \drop{} question to enforce consistency between the \find{} modules. For example, 
\question{How many \underline{percentage points} did the \underline{population of non-Hispanic Whites} drop from 1990 to \underline{2010}?} is paired with the generated question \question{What percentage of the population was non-Hispanic Whites in 2010?}.

\section{Experiments}
\label{sec:exp}
\paragraph{Dataset and Setup}
We perform experiments on the subset of the DROP dataset~\citep{DROPDua17} that is covered by the modules in Text-NMN. This subset is a union of the data used by \citet{GuptaNMN20} and the question decomposition annotations in the \qdmr{} dataset~\citep{break-wolfson2020}. All questions in our dataset contain program annotations (heuristically annotated by \citet{GuptaNMN20}; crowd-sourced in \qdmr{}). The program annotations only supervise the layouts of the modules, and not the intermediate outputs. We only use these programs for training; all validation and test results are based on predicted programs. 
Our complete subset of \drop{} contains 23215 question-answer pairs. For an i.i.d. split, since the \drop{} test set is hidden, we split the training set into train/validation and use the provided validation set as the test set. Our train / validation / test sets contain 18299 / 2460 / 2456 questions, respectively. 
In the training data, we found 7018 naturally-occurring pairings for 6039 questions (\S\ref{ssec:found-data}); construct template-based paired examples for 10882 questions~(\S\ref{sssec:construct-paired}); and generate 2632 questions paired with 2079 \drop{} questions~(\S\ref{sssec:generate-paired}). 
We use these paired examples to compute $\pairedloss$, which is added as an additional training objective on top of the standard training regime for Text-NMN
(see \S\ref{appendix:exp-details} for details).
Note that we do not add any additional (question, answer) pairs to the data, only new (unlabeled) questions.

\paragraph{Baselines}
As we are studying the impact of our new paired learning objective, our main point of comparison is a Text-NMN trained without that objective.
Though the focus of our work is improving learning in structured interpretable models, we also show results from a strong, reasonably comparable black-box model for \drop{}, \mtmsn{}~\citep{MTMSN19}, to better situate the relative performance of this class of models.

Experimental details are described in \S\ref{appendix:exp-details}. We release all our data and code publicly at \url{nitishgupta.github.com/nmn-drop}.

\begin{table}
\centering
\resizebox{0.5\textwidth}{!}{
\begin{tabular}{l cc cc}
\toprule
\bf \multirow{2}[3]{*}{Model} & \multicolumn{2}{c}{\bf dev} &  \multicolumn{2}{c}{\bf test} \\
\cmidrule(lr){2-3}  \cmidrule(lr){4-5}
& \bf F1  & \bf EM & \bf F1 & \bf EM \\
\midrule
\textsc{MTMSN}                      & 66.2  & 62.4 & 72.8  & 70.3   \\
\midrule
\nmn{} Baseline                     & 62.6  & 58.0 & 70.3  & 67.0   \\
\addlinespace[1mm]
\textsc{NMN} + \foundloss{}         & 66.0  & 61.5 & 71.0  & 67.8   \\
\textsc{NMN} + \temploss{}          & 66.2  & 61.4 & 72.3  & 69.2   \\
\textsc{NMN} + \qgenloss{}          & 63.7  & 58.9 & 71.2  & 68.4   \\
\addlinespace[1mm]
\textsc{NMN} + \allloss{}      & \textbf{66.3}  & \textbf{61.6} & \textbf{73.5}  & \textbf{70.5}   \\
\bottomrule
\end{tabular}
}
\centering
    \caption{ \label{tab:mainresults} \textbf{Performance on \drop{} (pruned):} Using our paired objective with all different kinds of paired-data leads to improvements in \nmn{}. Model achieves the best performance when all kinds of paired-data are used together.}
\end{table}

\begin{table*}[tbh]
\small
\centering
\captionsetup{font=normal}
\resizebox{1.0\textwidth}{!}{
\begin{tabular}{lccccccc}
\toprule
\multirow{2}[3]{*}{\textbf{Model}} & 
\multirow{2}[3]{*}{\begin{tabular}{@{}c@{}}\textbf{Performance} \\ (F$_1$ Score)\end{tabular}} & \multirow{2}[3]{*}{\begin{tabular}{@{}c@{}}\textbf{Overall Faithfulness} \\ (cross-entropy$^{*}$ $\downarrow$) \end{tabular}} & 
\multicolumn{5}{c}{Module-wise Faithfulness$^{*}$ ($\downarrow$)} \\
\cmidrule(lr){4-8}
&    &  & \module{find}  & \module{filter} & \module{num-date}$^\dagger$ & \module{project} & \module{min-max}$^\dagger$ \\
\midrule
\nmn{}                                  & 70.3  & 46.3      & 14.3  & 21.0   & 30.6  & 0.9   & 1.4       \\
\nmn{} + \allloss{}                     & 73.5  & 13.0      & 4.4   & 5.7    & 8.3   & 1.4   & 1.2       \\
\bottomrule
\end{tabular}
}
\caption{ 
\label{tab:faithful-results}
    \textbf{Faithfulness scores}: Using the paired objective significantly improves intermediate output predictions.
    {\footnotesize $^\dagger$denotes the average of \findnum{} \& \finddate{} and \findminnum{} \& \findmaxnum{}.}
}
\end{table*}

\subsection{In-distribution Performance}
\label{ssec:exp:iid}
We first evaluate the impact of our proposed paired objective on in-distribution generalization.
Table~\ref{tab:mainresults} shows the performance of the NMNs, trained with and without the paired objective, using different types of paired examples. We see that the paired objective always leads to improved performance; test F1 improves from $70.3$ F1 for the vanilla NMN to (a) $71$ F1 using naturally-occurring paired examples (\foundloss{}), (b) $72.3$ F1 using template-based paired examples (\temploss{}), and (c) $71.2$ F1 using model-generated paired examples (\qgenloss{}).
Further, the model achieves the best performance when all kinds of paired examples are combined, improving the performance to $73.5$ F1 (\allloss{}).\footnote{The improvement over the baseline is statistically significant (p = $0.01$) based on the Student's t-test. 
Test numbers are much higher than dev since the test set contains 5 answer annotations for each question.}
Our final model also outperforms the black-box \mtmsn{} model.

\subsection{Measuring faithfulness of NMN Execution}
\label{ssec:exp:faithful}
As observed by~\citet{subramanian-etal-2020-obtaining}, training a NMN only using the end-task supervision can lead to learned modules whose behaviour is \emph{unfaithful} to their intended reasoning operation, even when trained and evaluated with gold programs.
That is, even though the NMN might produce the correct final output, the outputs of the modules are not as expected according to the program (e.g., outputting only the longest field goal for the \find{}\modarg{field goal} execution), and this leads to markedly worse generalization on DROP. 
They release annotations for \drop{} containing the correct spans that should be output by each module in a program, and propose a cross-entropy-based metric to quantify the divergence between the output distribution over passage tokens and the annotated spans. 
A lower value of this metric denotes better faithfulness of the produced outputs. 

We evaluate whether the use of our paired objective to indirectly supervise latent decisions (module outputs) in a NMN indeed leads to more faithful execution. 
In Table~\ref{tab:faithful-results} we see that the NMN trained with the proposed paired objective greatly improves the overall faithfulness ($46.3 \rightarrow 13.0$) and also leads to huge improvements in most modules. 
This faithfulness evaluation shows that enforcing consistency between shared substructures provides the model with a dense enough training signal to learn correct module execution.
That is, not only does the model performance improve by using the paired objective, this faithfulness evaluation shows that the model's performance is improving \emph{for the right reasons}. In \S\ref{ssec:exp:analysis} we explore how this faithfulness is actually achieved.

\begin{table*}[tb]
\small
\centering
\begin{tabular}{l ccc ccc}
\toprule
\bf \multirow{2}[3]{*}{Model} &
\multicolumn{3}{c}{\bf Complex Arithmetic} &
\multicolumn{3}{c}{\bf Filter-ArgMax} \\
\cmidrule(lr){2-4}  \cmidrule(lr){5-7}
& {\bf dev}  & {\bf test w/o G.P.} & {\bf test w/ G.P.} & {\bf dev}  & {\bf test w/o G.P.} & {\bf test w/ G.P.} \\
\midrule
\textsc{MTMSN}              & 67.3  & \multicolumn{2}{c}{44.1}      & 67.5  &  \multicolumn{2}{c}{59.3}      \\ 
\midrule
\nmn{}                      & 64.3  & 29.5  & 42.1                  & 65.0  &  55.6  & 59.7      \\
\nmn{} + \allloss{}         & 67.2  & 47.2  & \textbf{54.7}         & 65.5  &  62.3  & \textbf{71.5}      \\
\bottomrule
\end{tabular}
\caption{ \textbf{Measuring compositional-generalization}: NMN performs substantially better when trained with the paired objective and performs even better when gold-programs are used for evaluation (w/ G.P).}
\label{tab:comp-split}
\end{table*}

\subsection{Evaluating Compositional Generalization}
\label{ssec:exp:compgen}
A natural expectation from structured models is that the explicit structure should help the model learn \textit{reusable} operations that generalize to novel contexts. 
We test this capability using the \emph{compositional generalization} setup of ~\citet{finegan-dollak-etal-2018-improving}, where the model is tested on questions whose program templates are unseen during training. 
In our case, this tests whether module executions generalize to new contexts in a program.

We create two test sets to measure our model's capability to generalize to such out-of-distribution examples. 
In both settings, we identify certain program templates to keep in a held-out test set, and use the remaining questions for training and validation purposes. 

\paragraph{Complex Arithmetic} 
This test set contains questions that require addition and subtraction operations in complex contexts: questions whose program contains \numadd{}/\numdiff{} as the root node, but the program is \emph{not} the simple addition or subtraction template \numadd{}/\numdiff{}(\findnum{}(\find{}), \findnum{}(\find{})). 
For example, \question{How many more mg/L is the highest amount of arsenic in drinking water linked to skin cancer risk than the lowest mg/L amount?}, with program
\numdiff{}(\findnum{}(\findmaxnum{}(\find{})), \findnum{}(\findminnum{}(\find{}))).

\paragraph{Filter-Argmax}
This test set contains questions that require an argmax operation after filter: programs that contain the subtree
\findmaxnum{}/\findminnum{}(\filter{}($\cdot$)).
For example, \question{Who scored the shortest touchdown in the first half?}, with program \project{}(\findmaxnum{}(\filter{}(\find{}))).

\paragraph{Performance}
In Table~\ref{tab:comp-split} we see that a NMN using our paired objective outperforms both the vanilla NMN and the black-box \mtmsn{} on both test sets.\footnote{The test set size is quite small, so while the w/ G.P. results are significantly better than \mtmsn{} (p~=~$0.05$), we can't completely rule out noise as the cause for w/o G.P. outperforming \mtmsn{} (p = $0.5$), based on the Student's t-test.}
This shows that enforcing consistent module behavior also improves their performance in novel contexts and as a result allows the model to generalize to out-of-distribution examples.
We see a further dramatic improvement when the model is evaluated using gold programs.  This is not surprising since it is known that semantic parsers (including the one in our model) often fail to generalize compositionally~\citep{finegan-dollak-etal-2018-improving, Lake2018GeneralizationWS, Bahdanau2019CLOSUREAS}.  Recent advancements in semantic parsing models that aim at compositional generalization should help improve overall model performance~\citep{Lake2019CompositionalGT, Korrel2019TranscodingCU, Herzig2020SpanbasedSP}.

\setlength\intextsep{1pt}
\begin{table}
\small
\centering
\resizebox{0.5\textwidth}{!}{
\begin{tabular}{l @{\hskip 4pt}c@{\hskip 4pt}c@{\hskip 4pt}c@{\hskip 4pt}c}
\toprule
\multirow{2}[2]{*}{\begin{tabular}{@{}l@{}}\textbf{Model} \end{tabular}} &
\multicolumn{3}{c}{\bf Test F1} &
\multirow{2}[2]{*}{\begin{tabular}{@{}c@{}} {\bf Faithful.-}\\{\bf score ($\downarrow$)} \end{tabular}} \\
\cmidrule(lr){2-4}
& 
Overall &
Min-Max &
Count &
 \\
\midrule
\nmn{}                                      & 57.4  & 82.1  & 36.2      & 110.4 \\
\, + $\mathcal{L}_{\text{max+min}}$         & 60.9  & \textbf{85.5}  & 39.7      & 56.5  \\
\, + $\mathcal{L}_{\text{max+count}}$       & 60.8  & 81.4  & 43.0      & 99.2  \\
\, + $\mathcal{L}_{\text{max+min+count}}$   & \textbf{71.1}  & 85.4  & \textbf{58.8}      & \textbf{25.9}  \\
\bottomrule
\end{tabular}
}
\caption{ 
\small
Using constructed paired examples for all three types of questions---min, max, and count---leads to dramatically better count performance. Without all three, the model finds shortcuts to satisfy the consistency constraint and does not learn correct module execution.
}
\label{tab:count-analysis}
\end{table}

\subsection{Analysis}
\label{ssec:exp:analysis}
We perform an analysis to understand how augmented paired examples---ones that do not contain end-task supervision---help in improving latent decision predictions. We conduct an experiment on a subset of the data containing only min, max and count type questions; programs in (1)-(6) from \S\ref{sssec:construct-paired}. 
We see a dramatic improvement over the baseline in count-type performance when paired examples for all three types of questions are used; answer-F1 improves from $36.2 \rightarrow 58.8$, and faithfulness from $110.4 \rightarrow 25.9$. This verifies that without additional supervision the model does indeed perform the min/max operation internal to its parameters and ground to the output event instead of performing the correct \find{} operation~(\S\ref{sec:paired-nmn}). As a result, the \find{} computation that \emph{should} be shared with the count questions is not actually shared, hurting performance.
By indirectly constraining the \find{} execution to produce consistent outputs for all three types of questions via the constructed question~(Fig.~\ref{fig:construct-paired}), the model learns to correctly execute \find{}, resulting in much better count performance. 
Using paired examples only for max and count questions ($\mathcal{L}_{\text{max+count}}$) does not constrain the \find{} operation sufficiently---the model has freedom to optimize the paired objective by learning to incorrectly ground to the max-event mention for both the original and constructed question's \find{} operation.
This analysis reveals that augmented paired examples are most useful when they form enough indirect connections between different types of instances to densely characterize the decision boundary around the latent decisions.

\section{Related Work}
The challenge in learning models for complex problems can be viewed as the emergence of artificially simple decision boundaries due to data sparsity and the presence of spurious dataset biases~\citep{Gardner2020EvaluatingNM}.  
To counter data sparsity, data augmentation techniques have been proposed to provide a compositional inductive bias to the model~\citep{Chen2020NeuralSR, Andreas2020GoodEnoughCD} or induce consistent outputs~\citep{LogicGuidedDA-Asai2020, ribeiro2019red}.
However, their applicability is limited to problems where the end-task supervision ($y$) for the augmented examples can be easily inferred. 
To counter dataset biases, model-based data pruning~\citep[AFLite;][]{Bras2020AdversarialFO} and subsampling~\citep{Oren2020-semparse} have been proposed. All the techniques above modify the training-data distribution to remove a model's propensity to find artificially simple decision boundaries, whereas we modify the training objective to try to accomplish the same goal. 
Ensemble-based training methodology~\citep{Clark2019DontTT, Stacey2020ThereIS} has been proposed to learn models robust to dataset artifacts; however, they require prior knowledge about the kind of artifacts present in the data.

Our approach, in spirit, is related to a large body of work on learning structured latent variable models.
For example, prior work has incorporated indirect supervision via constraints~\citep{Graa2007ExpectationMA, chang2007guiding, ganchev2010posterior} or used negative examples with implausible latent structures~\citep{ContrastiveET-Smith2005, Chang2010StructuredOL}.
These approaches use auxiliary objectives on a single training instance or global conditions on posterior distributions, whereas our training objective uses \emph{paired examples}.

\section{Conclusion}
We propose a method to leverage \emph{paired examples}---instances that share internal substructure---to provide a richer training signal to latent decisions in compositional model architectures. 
We provide a general formulation of this technique which should be applicable to a broad range of models. To validate this technique, we present a case study on text-based neural module networks, showing how to apply the general formulation to a specific task.
We explore three methods to acquire paired examples and empirically show that our approach leads to substantially better in- and out-of-distribution generalization of a neural module network in complex compositional question answering.  We also show that using our paired objective leads to improved prediction of latent decisions.

A lot of recent work is exploring the use of closely related instances for improved evaluation and training.  Ours is one of the first works to show substantial improvements by modifying the training objective to try to make better use of the local decision surface.  These results should encourage more work exploring this direction.

\bibliography{references}
\bibliographystyle{acl_natbib}  

\appendix
\section{Appendix}

\subsection{Experimental Details}
\label{appendix:exp-details}
All models use the bert-base-uncased model to compute the question and passage contextualized representations. 
For all experiments (including all baselines), we train two versions of the model with different seed values, and choose the one that results in higher validation performance. All models are trained for a maximum number of $40$ epochs, with early stopping if validation F1 does not improve for $10$ consecutive epochs.
For \mtmsn{}, we use the hyperparameters provided with the original code. 
For \nmn{}, we use a batch size of $2$ (constrained by a $12$GB GPU) and a learning rate of \texttt{1e-5}.
The question parser in the Text-NMN uses a 100-dimensional, single-layer LSTM decoder.
Our code is written using the AllenNLP library~\citep{Gardner2017AllenNLP}.

Training objective: We simply add the paired objective $\pairedloss{}$ to the training objective of Text-NMN~\citep{GuptaNMN20} which includes a maximum likelihood objective to predict the gold program, a maximum likelihood objective for gold answer prediction from the program execution, and an unsupervised auxiliary loss to aid information extraction. We do not use any heuristically-obtained intermediate module output supervision used in \citet{GuptaNMN20}.

\paragraph{Dataset}
As mentioned in \S\ref{sec:exp}, our dataset is composed of two subsets of \drop{}: (1) the subset of \drop{} used in \citet{GuptaNMN20}---this contains 3881 passages and 19204 questions---and (2) question-decomposition meaning representation (QDMR) annotations from \qdmr{}~\citep{break-wolfson2020}---this contains 2756 passages with a total of 4762 questions. After removing duplicate questions we are left with 23215 questions in total.
We convert the program annotations in QDMR to programs that conform to the grammar induced by the modules in Text-NMN using simple transformations.

We will publicly release all our code, data, and trained-model checkpoints for reproducibility.

\subsection{Model-generated Paired Examples}
\label{appendix:modgen}
For the question generation model we use the BART-large model and train for $1$ epoch using a learning rate of \texttt{3e-5}. From the SQuAD dataset, we use as training data only questions whose answer text appears exactly once in the passage.

As mentioned in \S\ref{sssec:generate-paired}, we only generate paired questions for non-football questions, based on two frequent observations, both related to domain shift. 
Consider this snippet from a passage containing a football game summary:

\emph{
Following their road loss to the Steelers, the Browns flew to M\&T Bank Stadium for an AFC North rematch with the Baltimore Ravens. .....   the Browns showed signs of life as QB Derek Anderson completed a 3-yard TD pass to WR Joe Jurevicius. .....  Cleveland tied the game at 17-17 with Anderson's 14-yard TD pass to WR Braylon Edwards. ..... the Ravens took over for the rest of the game with Boller's 77-yard TD pass
 to WR Demetrius Williams. 
}

If we generate a question conditioned on the number 3 as the answer, our QG model typically generates a question such as \textit{How many yards was the TD pass?} or \textit{How many yards Anderson's pass?}. Both of these questions have \emph{incorrect presuppositions}, as the answer to them is not just the conditioned answer (3)---there are multiple possible answers in the given paragraph. In the football game summary domain, it is common for a single event type to contain multiple mentions like this, which our QG model trained on SQuAD cannot handle.
Similarly, we observe that the QG model generates nonsensical questions for numbers associated to game scores (e.g. 17-17), likely due to domain shift. Future work should look into QG models that can operate under different domains to generate paired examples.

\end{document}